\documentclass[letterpaper, 10 pt, conference]{ieeeconf}  

\usepackage{amsmath}
\usepackage{amssymb}
\usepackage{graphicx}
\usepackage{booktabs}
\usepackage{multirow} 
\usepackage{graphicx}
\usepackage{subcaption}
\usepackage{hyperref}

\IEEEoverridecommandlockouts                              

\title{\LARGE \bf
ParkFormer: A Transformer-Based Parking Policy with Goal Embedding and Pedestrian-Aware Control
}

\author{Jun Fu$^{1}$, Bin Tian$^{2,*}$, Haonan Chen$^{3}$, Shi Meng$^{4}$, Tingting Yao$^{5}$
\thanks{*This work was supported by the Key-Area Research and Development Program of Guangdong Province (2020B0909050001), the National Key Research and Development Program of China (2022YFB4703700), the Key Research and Development Program of Shaanxi Province (2024CY2-GJHX-49).}
\thanks{$^{1}$Jun Fu, $^{3}$Haonan Chen, $^{4}$Shi Meng and $^{5}$Tingting Yao are with Institute of Automation, Chinese Academy of Sciences,Beijing 100190, China (e-mail: fujun2023@ia.ac.cn; chenhaonan2024@ia.ac.cn; mengshi2022@ia.ac.cn; tingting.yao@ia.ac.cn)}%
\thanks{$^{*2}$Bin Tian is with Institute of Automation, Chinese Academy of Sciences,Beijing 100190, China, also with the School of Artificial Intelligence, University of Chinese Academy of Sciences, Beijing 100049, China (e-mail: bin.tian@ia.ac.cn).}%
}

\begin{document}

\maketitle
\thispagestyle{empty}
\pagestyle{empty}

\begin{abstract}
Autonomous parking plays a vital role in intelligent vehicle systems, particularly in constrained urban environments where high-precision control is required. While traditional rule-based parking systems struggle with environmental uncertainties and lack adaptability in crowded or dynamic scenes, human drivers demonstrate the ability to park intuitively without explicit modeling. Inspired by this observation, we propose a Transformer-based end-to-end framework for autonomous parking that learns from expert demonstrations. The network takes as input surround-view camera images, goal-point representations, ego vehicle motion, and pedestrian trajectories. It outputs discrete control sequences including throttle, braking, steering, and gear selection. A novel cross-attention module integrates BEV features with target points, and a GRU-based pedestrian predictor enhances safety by modeling dynamic obstacles. We validate our method on the CARLA 0.9.14 simulator in both vertical and parallel parking scenarios. Experiments show our model achieves a high success rate of 96.57\%, with average positional and orientation errors of 0.21 meters and 0.41 degrees, respectively. The ablation studies further demonstrate the effectiveness of key modules such as pedestrian prediction and goal-point attention fusion. The code and dataset will be released at: \url {https://github.com/little-snail-f/ParkFormer}.

\end{abstract}

\section{INTRODUCTION}
With the rapid advancement of autonomous driving technologies, autonomous parking has emerged as a critical research topic due to its emphasis on low-speed and high-precision maneuvers. Typical parking scenarios involve narrow lanes, densely distributed obstacles, and frequent dynamic disturbances, which introduce substantial challenges to perception, trajectory planning, and fine-grained motion control.
\begin{figure}[t]
    \centering
    \includegraphics[width=\linewidth]{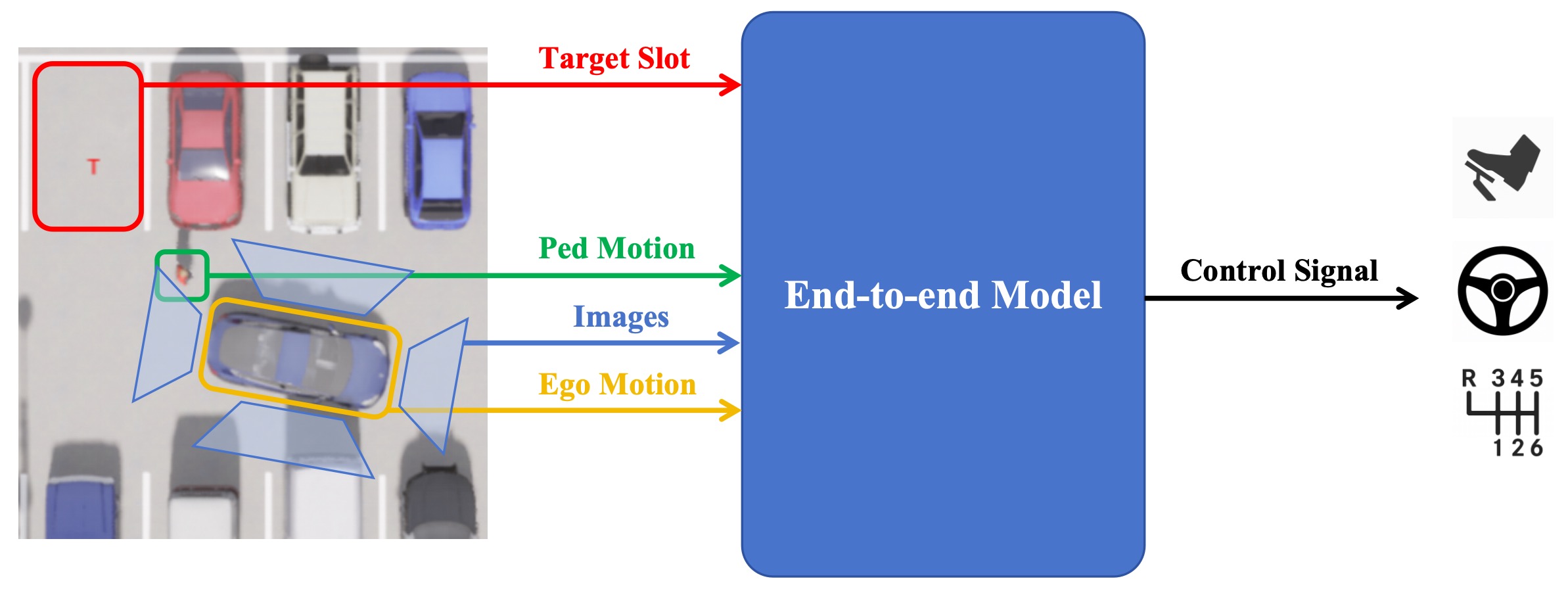}  
    \caption{Overview of the proposed framework. The system comprises distinct yet interactive modules for visual perception, pedestrian trajectory prediction, goal encoding, and dynamic planning, enabling safe and precise parking in complex environments.}
    \label{fig:architecture}
\end{figure}
Conventional parking systems typically adopt a multi-stage pipeline architecture that decouples the system into independent modules for perception, mapping, prediction, planning, and control \cite{c1}. Although this modular design offers advantages in terms of interpretability and engineering flexibility, it often suffers from error accumulation and interface coupling issues, especially in complex and unpredictable real-world environments. These issues degrade system robustness and limit scalability in real-world deployments.
In contrast, human drivers are able to perform parking maneuvers by relying on intuitive understanding and driving experience, without the need for precise maps or detailed planning. This observation has inspired the adoption of end-to-end learning paradigms for autonomous parking, enabling the system to directly learn driving policies from expert demonstrations in a data-driven manner.

Despite advances in end-to-end autonomous driving methods that successfully map raw sensor data directly to control outputs, most existing works primarily address open-road scenarios. The application of these methods to dynamic parking environments remains limited due to specific challenges, including:
(1) The presence of heterogeneous parking slot types, such as perpendicular and parallel parking, makes policy generalization difficult;
(2) Ensuring safe and efficient interactions with pedestrians and dynamic obstacles, which demands accurate predictive modeling of their trajectories and behaviors.

To address the aforementioned challenges, this paper proposes a vision-driven end-to-end framework for dynamic autonomous parking. Specifically, we design a multi-task perception network based on surround-view fisheye cameras, which performs joint perception of parking slot topology, vehicle pose, and pedestrian intent through a spatio-temporal feature fusion mechanism.
On top of this, we introduce a goal point encoder that maps heterogeneous parking slot geometries into a shared low-dimensional latent space, enabling unified policy modeling for both perpendicular and parallel parking scenarios.
To evaluate the effectiveness of our approach, we construct a dynamic parking dataset on the CARLA simulation platform and conduct extensive experiments to validate the system performance.

Overall, the main contributions of this work include:
\begin{itemize}
\item We propose an end-to-end autonomous parking model tailored for dynamic environments, which enables the vehicle to efficiently and accurately park in the target slot while safely interacting with pedestrians.
\item We introduce a novel goal-point fusion mechanism that encodes heterogeneous parking slot geometries and supports a unified policy for both perpendicular and parallel parking scenarios.
\item We release a simulation-based parking dataset with enhanced scene diversity and dynamic pedestrian agents, establishing a more realistic benchmark for end-to-end parking research.
\end{itemize}

\section{LITERATURE REVIEW}

In recent years, the end-to-end paradigm has achieved remarkable progress in autonomous driving, driven largely by the adoption of advanced neural architectures such as Transformers. This section reviews representative works categorized into two primary application domains: urban driving scenarios and parking scenarios.
\subsection{Urban Driving Scenarios}
The end-to-end autonomous driving methods generally utilize inputs from cameras, LiDAR, and multi-modal sensor fusion to efficiently bridge perception and decision-making. Early studies, such as CIL \cite{cil}, pioneered the direct mapping from front-view images, current vehicle states, and high-level commands to control signals. However, their performance was limited under complex driving conditions such as sharp turns. To improve the robustness and predictive capabilities of driving decisions, TCP \cite{tcp} introduced a trajectory-guided attention mechanism, enabling multi-step control based on trajectory forecasting. Further, Pix2Planning \cite{Pix2Planning} explicitly employed a Transformer decoder by concatenating the target location and historical trajectory as input, enabling fine-grained trajectory predictions. 

Transformer-based architectures \cite{attention} have further expanded the modeling capabilities for urban driving. Wayformer~\cite{Wayformer} proposed a Transformer-based multi-agent trajectory prediction model with strong spatiotemporal interaction modeling capabilities. BEVFormer~\cite{BEVFormer} integrated Transformer modules with BEV representations, enabling efficient feature transformation from multi-view images to BEV space and significantly improving perception accuracy. BEVFusion~\cite{BEVFusion} built a unified Transformer-based perception framework to achieve multi-modal fusion and prediction in BEV space. The VAD series~\cite{Vad} leveraged semantically guided Transformer encoders for multi-agent behavior modeling, and VADv2~\cite{Vad2} introduced trajectory distribution modeling and cascaded Transformer decoders to enhance end-to-end planning under complex scenarios and uncertainty. FSGA~\cite{FSGA}, based on the HiVT~\cite{Hivt} framework, incorporated future scene graph mechanisms and graph attention to explicitly model agent-lane interactions. RCTrans~\cite{Rctrans} employed a Transformer query mechanism and sequential decoding structure for radar-camera fusion, effectively improving cross-modal alignment and 3D detection robustness under sparse radar conditions.

To address diverse driving scenarios, researchers have proposed specialized Transformer designs. SwapTransformer \cite{SwapTransformer} alternates attention between temporal and feature dimensions to improve highway overtaking and lane changes. DualAT \cite{DualAT} adopts a dual-attention mechanism to fuse image and BEV features, with multi-task supervision for enhanced urban robustness. MAGNet \cite{Multi-Task} combines trajectory and control prediction using a gating network for adaptive fusion, improving generalization across driving contexts.

Recently, several studies have focused on integrating perception, prediction, and planning within end-to-end frameworks to improve overall system performance. UniAD \cite{uniad} proposes a unified architecture for joint multi-task optimization. GenAD \cite{GenAD} treats trajectory prediction and planning as a unified generative process in a latent space, explicitly modeling interactions and priors to achieve efficient and accurate planning. ThinkTwice \cite{Think twice} introduces a scalable "Look-Predict-Refine" decoder structure that progressively refines decisions, enhancing the safety and controllability of driving behaviors.

\begin{figure*}[htbp]
    \centering
    \includegraphics[width=0.95\textwidth]{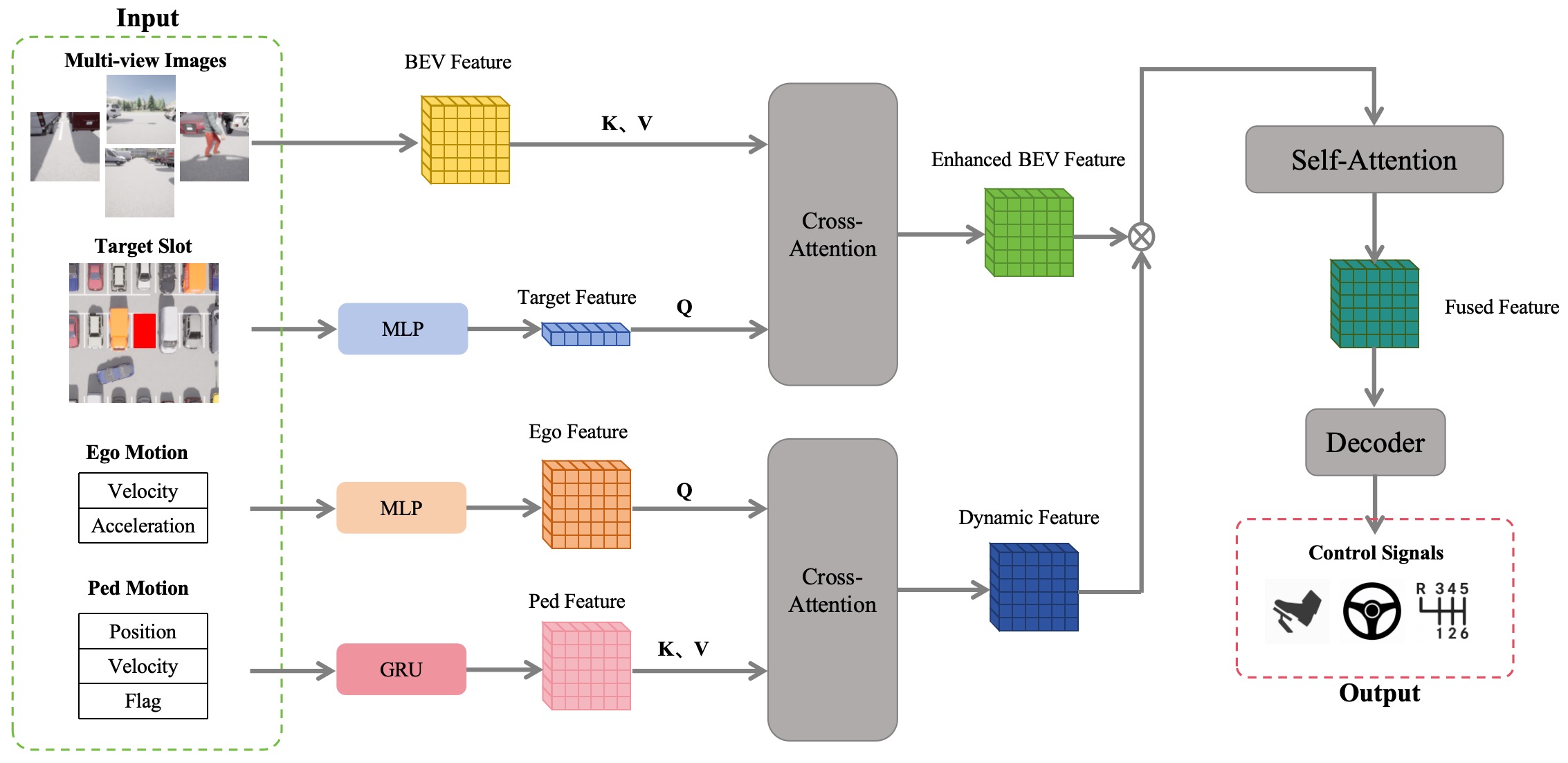}
    \caption{An overview of the proposed Transformer-based end-to-end framework for autonomous parking in dynamic environments. The model processes multi-view RGB images, the target parking slot information, ego-motion states, and pedestrian trajectories as inputs. It consists of four main components: (1) an image and goal encoder that generates goal-conditioned BEV features; (2) an ego-pedestrian encoder that models dynamic contextual information; (3) a multimodal feature fusion module that aggregates spatial and dynamic cues; and (4) a control decoder that autoregressively predicts discrete control commands over multiple future steps.}
    \label{overview}
\end{figure*}

\subsection{Parking Scenarios}
Al-Mousa et al. \cite{park} proposed a reinforcement learning-based modular reverse parking method that divides the parking task into three independently trained stages, employing top-view images for environmental perception. However, this approach relies on fixed initial positions and parking layouts, limiting its generalization capabilities. Selvaraj et al. \cite{v8} integrated deep reinforcement learning with YOLOv8 object detection, utilizing real-time heatmap guidance for parking decisions within the CARLA simulation environment, and introduced voice interaction and a privacy-oriented Stealth Parking mode, thereby enhancing system adaptability and practicality. RL-OGM-Parking \cite{RL-OGM-Parking} further presents a hybrid planning framework that combines a rule-based global planner with a reinforcement learning-based local controller, using LiDAR-generated occupancy grid maps (OGMs) as unified input. This method effectively bridges the perception gap between simulation and real-world deployment, demonstrating strong generalization and safety across complex scenarios.

On the other hand, end-to-end trajectory prediction methods have also been widely applied to autonomous parking tasks. The ParkPredict series \cite{ParkPredict}\cite{ParkPredict2} initially adopted CNN-LSTM and subsequently replaced it with Transformer architectures for vehicle trajectory prediction, while introducing a CNN-based intention prediction module to further improve trajectory prediction accuracy. E2E Parking \cite{E2E Parking} employed a Transformer to directly map images and motion states to parking control signals. ParkingE2E \cite{ParkingE2E} further extended this framework by convolving a binary map of the target parking slot into a query vector, subsequently fused with BEV feature maps to accurately predict target coordinates, demonstrating strong performance through real-world validation. TransParking \cite{TransParking} further introduced a dual-decoder Transformer structure to separately model interactions of trajectory coordinates (X and Y), coupled with a Gaussian-kernel-based soft localization mechanism, achieving refined and robust trajectory predictions.

\section{METHODOLOGY}
\subsection{Problem Formulation}
To safely and precisely perform autonomous parking maneuvers in dynamic environments involving pedestrian interactions, we formulate the parking task as an end-to-end supervised learning problem.  The objective is to learn a policy directly mapping multimodal sensory inputs to vehicle control outputs.  Specifically, the collected training dataset $\mathcal{D}$ includes multiple trajectories, each defined as follows:

\begin{equation}
\mathcal{D} = \left\{ I_{i,j}^{k},\ G_{i,j},\ \mathbf{E}_{i,j},\ \mathbf{P}_{i,j},\ C_{i,j} \right\},
\end{equation}

where each element is defined explicitly as:
\begin{itemize}
\item $I_{i,j}^{k}$: RGB images captured by surround-view camera $k$ at timestep $j$ in the $i$-th trajectory.
\item $G_{i,j} = (x_g, y_g, \psi_g)$: Target parking slot represented by its 2D center coordinates and orientation angle.
\item $\mathbf{E}_{i,j} = [v_{i,j}, a_{i,j}]$: Ego-vehicle state, including velocity $v_{i,j}$ and acceleration $a_{i,j}$.
\item $\mathbf{P}_{i,j} = {\mathbf{P}^{\text{pos}}_{i,j}, \mathbf{P}^{\text{vel}}_{i,j}, \mathbf{P}^{\text{acc}}_{i,j}, \mathbf{P}^{\text{mask}}_{i,j}}$: Pedestrian dynamics containing positions, velocities, accelerations, and a binary existence mask.
\item $C{i,j}$: Expert-generated discrete control commands (acceleration, steering, gear) at timestep $j$.
\end{itemize}

The ultimate training objective is to learn a control policy $\pi_\theta$ that predicts control outputs based on the current observations. The training process is conducted by minimizing the following loss function:
\begin{equation}
\theta^* = \arg\min_{\theta} \ \mathbb{E}_{(I, G, \mathbf{E}, \mathbf{P}, C) \sim \mathcal{D}} \left[ \mathcal{L}\left(C,\ \pi_{\theta}(I, G, \mathbf{E}, \mathbf{P})\right) \right],
\end{equation}
where $\mathcal{L}$ denotes the control loss function, and $\pi_\theta$ represents the control policy network parameterized by $\theta$.

\subsection{Overview of Architecture}
Distinct from prior works, our architecture unifies dynamic context modeling and goal-conditioned spatial reasoning within a Transformer-based framework. As illustrated in Fig.~\ref{overview}, the model takes multi-view images, ego-motion states, pedestrian-motion states, and target parking slot information as input to predict multi-step control outputs. Key innovations include a pedestrian encoder for proactive dynamic modeling and a goal-conditioned cross-attention mechanism for spatial alignment. These designs enable joint reasoning over dynamic elements and spatial semantics, enhancing safety, precision, and generalization across complex parking scenarios.

\subsection{Perception Module: Image and Goal Encoder}
The perception module aims to construct a structured BEV representation that explicitly integrates the parking target semantics. Following the LSS method~\cite{lss}, consistent with the approach in \cite{E2E Parking}, multi-view surround images $I$ are first encoded into feature maps $F_{\text{img}}$ using a shared convolutional backbone (EfficientNet-B4). Each image feature is processed to predict a discrete depth distribution $D$, which is then fused through frustum construction to generate 3D feature volumes $F_{\text{frus}}$. These volumes are projected into a BEV voxel grid using the known camera intrinsics $K$ and extrinsics $T$, aggregating multi-view information into an initial BEV feature map $F_{\text{bev}}$.
To enhance the spatial reasoning capability of $F_{\text{bev}}$, we inject a two-dimensional sine-cosine positional encoding, ensuring that geometric consistency is preserved across the BEV grid.

Beyond constructing a general BEV map, our model explicitly incorporates the goal slot information to guide spatial feature extraction. The target parking slot state $G = (x_g, y_g, \psi_g)$, representing its center position and orientation, is projected through an MLP to obtain a dense goal embedding $T_g$. This embedding is used as the Query input to a multi-head cross-attention module, while the BEV features serve as both the Key and Value.

Through this goal-conditioned attention mechanism, the model selectively emphasizes BEV regions relevant to reaching the specified parking goal. The attention output yields a refined, goal-aware BEV token sequence $F_{\text{att}}$, effectively blending visual and semantic cues.

\subsection{Motion Context Encoding: Ego-Pedestrian Encoder}
To explicitly account for dynamic obstacles during parking maneuvers, we introduce an ego-pedestrian encoding module that jointly models the future motion of surrounding pedestrians and the current state of the ego-vehicle.

Specifically, a lightweight GRU-based encoder-decoder architecture is employed to capture the temporal evolution of pedestrian behaviors. At each timestep, the historical pedestrian states—including 2D positions, velocities, accelerations, and a binary existence mask are processed to predict future pedestrian trajectories over a short prediction horizon of 10 steps. To mitigate the accumulation of prediction errors during sequence modeling, teacher forcing is applied during training, and a linear projection is used to normalize the ground-truth trajectory dimensions, facilitating stable learning. In parallel, the ego-vehicle motion state, characterized by instantaneous velocity and acceleration, is encoded via a MLP into a compact latent feature. To retain temporal ordering information, a sine-cosine positional encoding is subsequently added.

Rather than treating pedestrian and ego features independently, we explicitly design a cross-attention mechanism to fuse these two modalities. The ego-motion token serves as the Query, while the predicted pedestrian embeddings, enriched with positional encodings, are used as the Keys and Values. Through this selective attention operation, the ego-vehicle dynamically focuses on the most relevant moving agents within its vicinity, enabling context-aware decision-making.

The resulting motion context embedding captures both the ego state and the surrounding pedestrian dynamics in a temporally consistent manner, and is propagated forward to the control prediction module.

\subsection{Feature Fusion and Control Decoder}

After obtaining the goal-aware BEV features and the dynamic motion context encoding, we perform multimodal feature fusion to jointly model spatial semantics and dynamic interactions. Specifically, the BEV token sequence $F_{\text{att}}$ and the motion context embeddings are concatenated along the feature dimension, forming a unified representation $F_{\text{fuse}}$. To preserve the sequential structure and relative spatial-temporal ordering, learnable positional embeddings are added to the fused sequence.

This integrated representation is then processed by a Transformer encoder consisting of multiple self-attention layers. The encoder captures high-order dependencies across different modalities and propagates task-critical information, such as goal proximity and pedestrian movement trends, throughout the latent space. This multimodal fusion enables the model to reason jointly over environmental context and dynamic elements when generating future actions.

For control prediction, we formulate the task as an autoregressive sequence modeling problem. Instead of directly regressing continuous control outputs, we discretize each control component—acceleration, steering angle, and gear state—into finite token spaces through adaptive binning strategies optimized for vehicle dynamics. The control sequence is initialized with a special Begin-of-Sequence (BOS) token and terminated with an End-of-Sequence (EOS) token, defined as:
\begin{equation}
S = [\text{BOS},\ c_0,\ c_1,\ \dots,\ c_T,\ \text{EOS}],\quad c_t = \{a_t, s_t, g_t\},
\end{equation}
where $a_t$, $s_t$, and $g_t$ are the discretized control tokens at time step $t$.

A Transformer decoder conditioned on $F_{\text{fuse}}$ generates the control sequence token-by-token in an autoregressive manner. During each decoding step, the previously generated tokens, fused environmental features, and positional encodings collectively inform the prediction of the next action. This design enforces strong temporal coherence across predicted actions, ensuring smooth, accurate, and collision-free parking maneuvers.

Finally, the decoder outputs are projected through a linear classification head to produce token-level probability distributions. The entire model is optimized by minimizing the sequential cross-entropy loss between predicted control sequences $\hat{S}$ and the ground-truth sequences $S_{\text{gt}}$:
\begin{equation}
\mathcal{L}_{\text{ctrl}} = \text{CrossEntropy}(\hat{S}, S_{\text{gt}}).
\end{equation}

\section{EXPERIMENTS}
\begin{figure}[t]
    \centering
    \subfloat[Town04-Opt: vertical parking scenario.]{
        \includegraphics[width=0.8\linewidth]{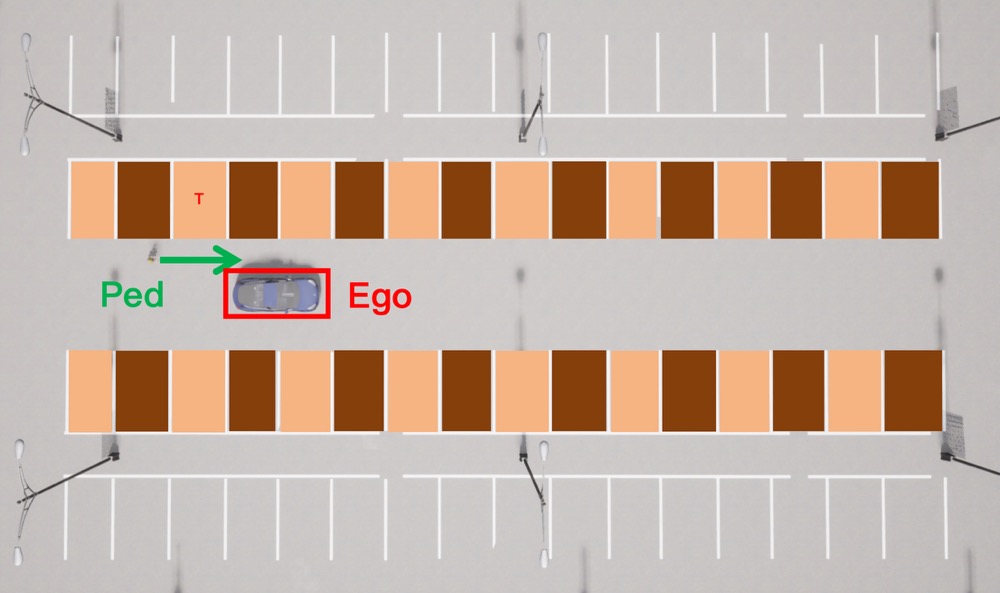}
        \label{fig:town04}
    } \\
    \vspace{0.5em}
    \subfloat[Town10HD-Opt: parallel parking scenario.]{
        \includegraphics[width=0.8\linewidth]{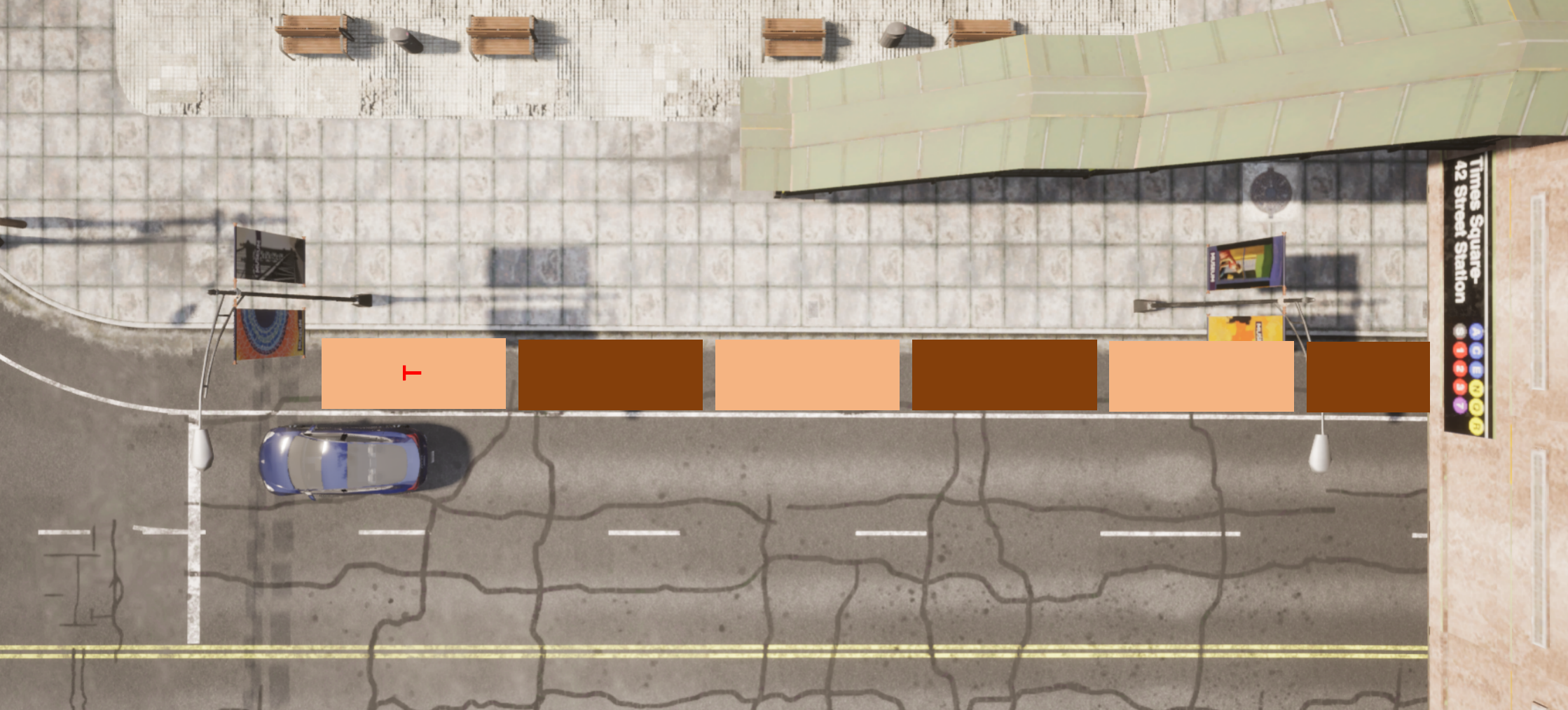}
        \label{fig:town10}
    }
    \caption{Top-down BEV views of two simulated parking scenarios in CARLA: (a) Town04-Opt (vertical layout), and (b) Town10HD-Opt (parallel layout). The target slot is marked with a red “T” and kept vacant, while other slots are randomly occupied.}
    \label{fig:parking_scenes}
\end{figure}
\subsection{Dataset Collection}
To facilitate robust training and evaluation, we design an enhanced data collection pipeline in the CARLA simulation environment, building upon the framework introduced in~\cite{E2E Parking}. Compared to the original setup, our pipeline significantly improves scenario diversity by incorporating both perpendicular and parallel parking layouts, along with randomized pedestrian dynamics. These enhancements result in a more realistic and challenging benchmark for end-to-end autonomous parking systems.

As shown in Fig.~\ref{fig:parking_scenes}, we design two distinct parking configurations: (1) a perpendicular parking lot with four rows of slots (Fig.~\ref{fig:town04}), and (2) a parallel parking layout organized along a single street side (Fig.~\ref{fig:town10}). The target slot, marked by a red “T”, is kept unoccupied, while other slots are randomly filled with static vehicles. Default CARLA objects are removed to ensure scene controllability.

In each episode, the ego vehicle is randomly initialized outside the target slot with lane-aligned heading.  A parking attempt is triggered within 7 meters of the target.  To simulate realistic dynamics, pedestrians with randomized motion are spawned.

Both manual driving and rule-based controllers are used during collection. The system synchronously records multimodal observations, including RGB images, depth maps, ego-motion states (velocity and acceleration), pedestrian trajectories, and control commands. Ground-truth BEV maps are generated by projecting 3D bounding boxes, and depth images are captured using front-mounted cameras aligned with the RGB view.

We retain only those trajectories satisfying strict quality constraints, i.e., final position and yaw errors below 0.5 meters and 0.5 degrees, respectively.

In total, 272 parking episodes were collected, comprising approximately 46,400 frames at 10 Hz. The dataset comprehensively covers both perpendicular and parallel parking scenarios and has been publicly released to promote reproducible research in autonomous parking systems.

\subsection{Implementation Details}
We develop a multi-scenario autonomous parking framework on the CARLA 0.9.14 simulation platform, covering two representative environments: Town04-Opt with 64 vertical parking slots (Fig.~\ref{fig:town04}) and Town10HD-Opt with 6 parallel parking slots (Fig.~\ref{fig:town10}).

The BEV representation spans a spatial range of $[-10\,\mathrm{m}, 10\,\mathrm{m}]$ along both $x$ and $y$ axes, discretized into a $200\times200$ grid at a $0.1\,\mathrm{m}$ resolution. A pretrained EfficientNet-B4 backbone extracts semantic features from multi-view images.

Our network adopts a Transformer-based architecture, with a feature fusion encoder and a control decoder, each comprising 4 layers and 8 attention heads, supplemented by sinusoidal positional encodings.

Training is conducted using PyTorch on an NVIDIA A40 GPU, with a batch size of 32 over 150 epochs. Approximately 45,000 multimodal frames sampled at 10 Hz are used, including RGB images, depth maps, ego states, and pedestrian states. The entire training process takes about 84 hours.

\begin{figure}[t]
    \centering
        \begin{subfigure}[b]{0.24\linewidth}
        \centering
        \includegraphics[width=\linewidth]{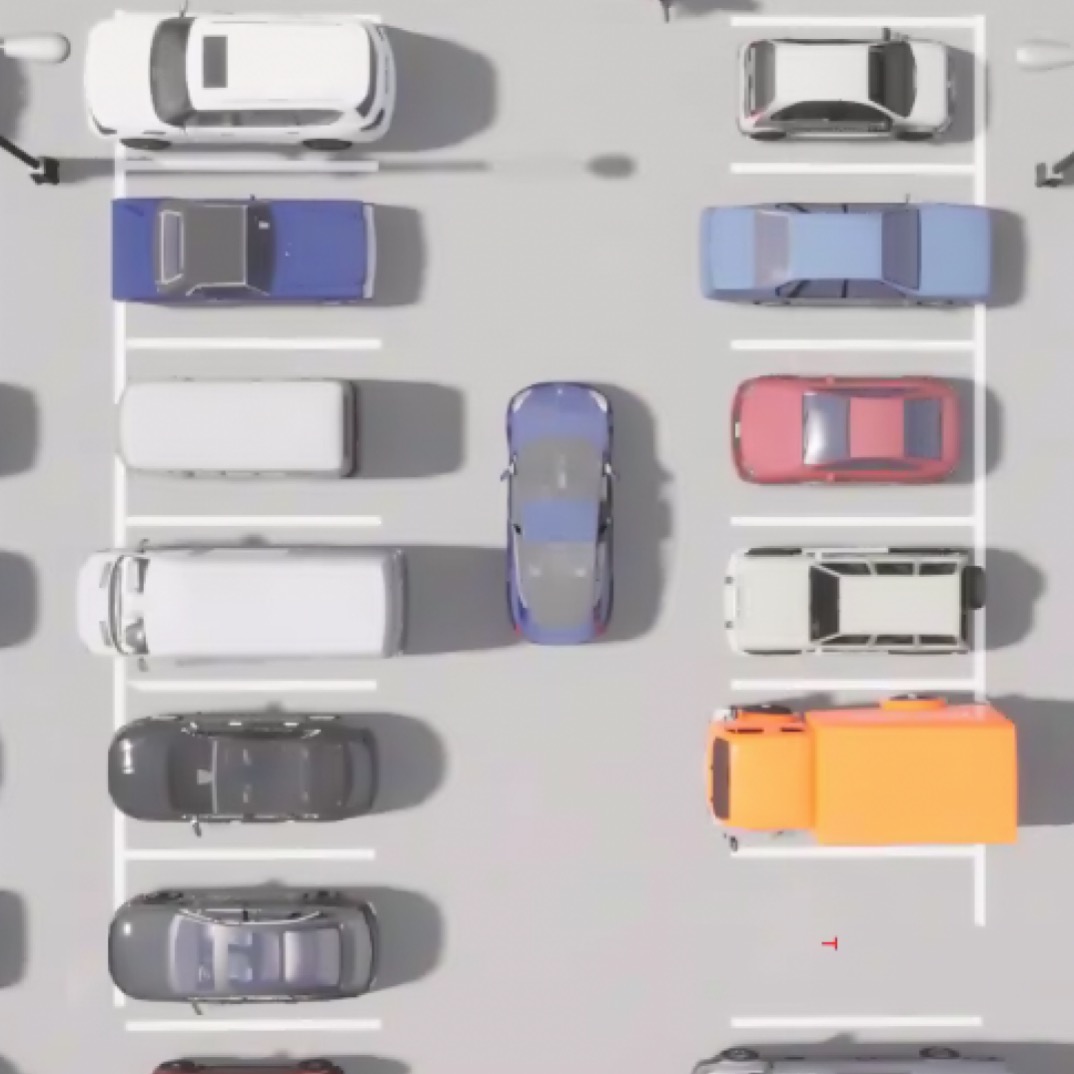}
        \caption{}
        \label{fig:bev1}
    \end{subfigure}
    \hfill
    \begin{subfigure}[b]{0.24\linewidth}
        \centering
        \includegraphics[width=\linewidth]{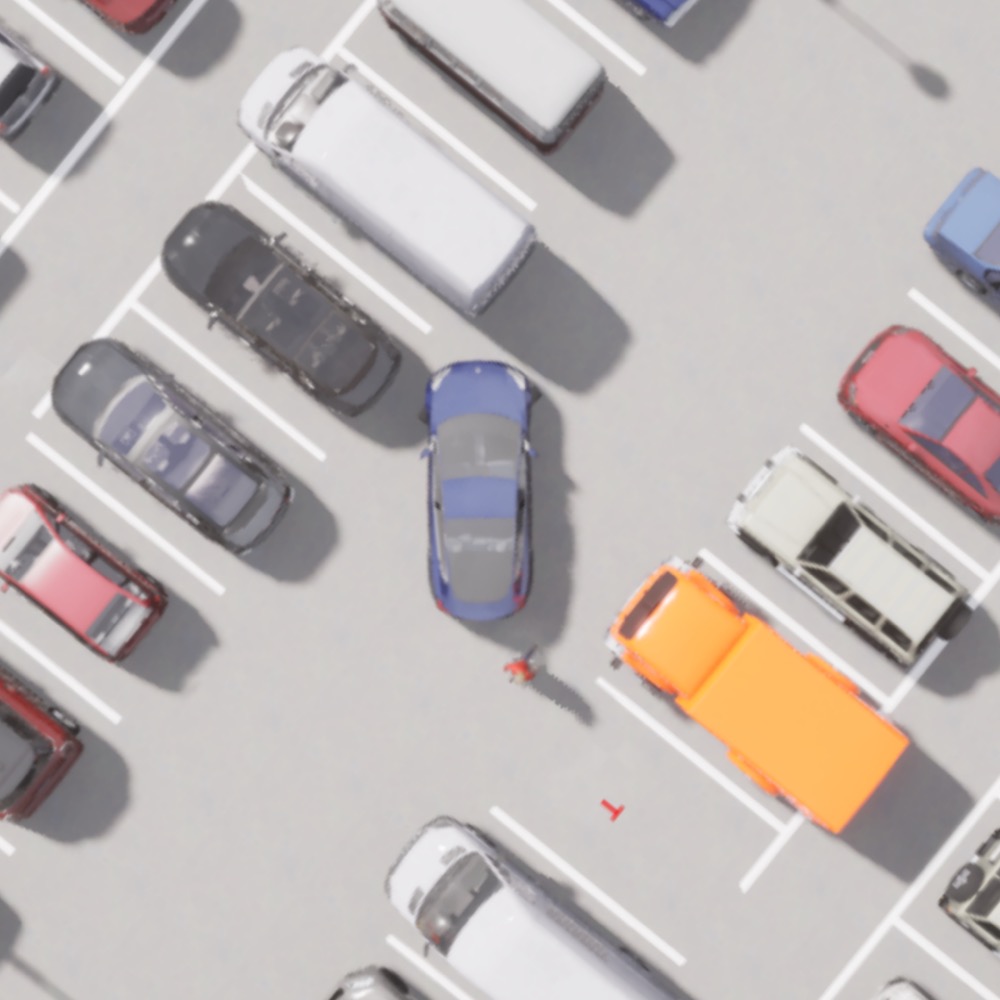}
        \caption{}
        \label{fig:bev2}
    \end{subfigure}
    \hfill
    \begin{subfigure}[b]{0.24\linewidth}
        \centering
        \includegraphics[width=\linewidth]{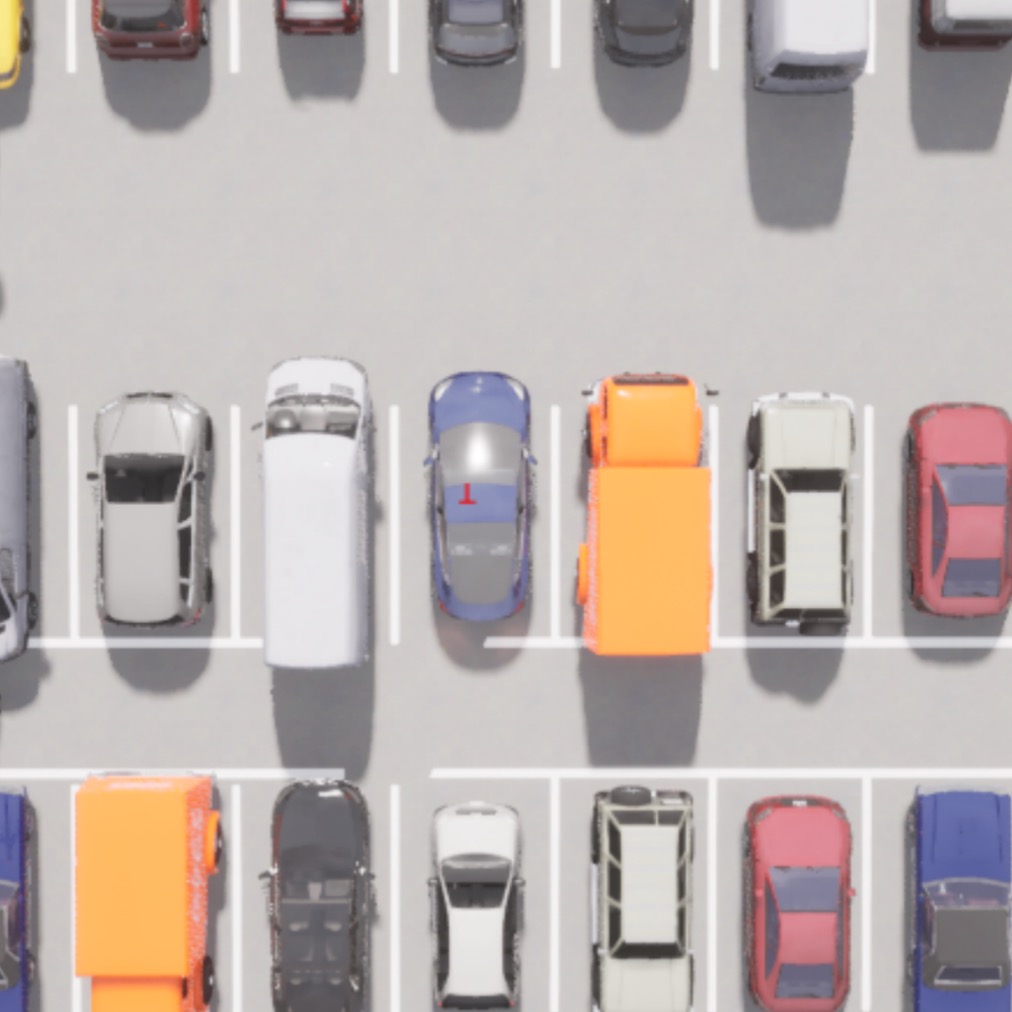}
        \caption{}
        \label{fig:bev3}
    \end{subfigure}
    \hfill
    \begin{subfigure}[b]{0.24\linewidth}
        \centering
        \includegraphics[width=\linewidth]{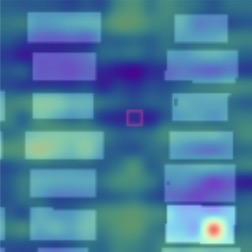}
        \caption{}
        \label{fig:att}
    \end{subfigure}

    \vspace{0.4em} 
 
    \begin{subfigure}[b]{0.24\linewidth}
        \centering
        \includegraphics[width=\linewidth]{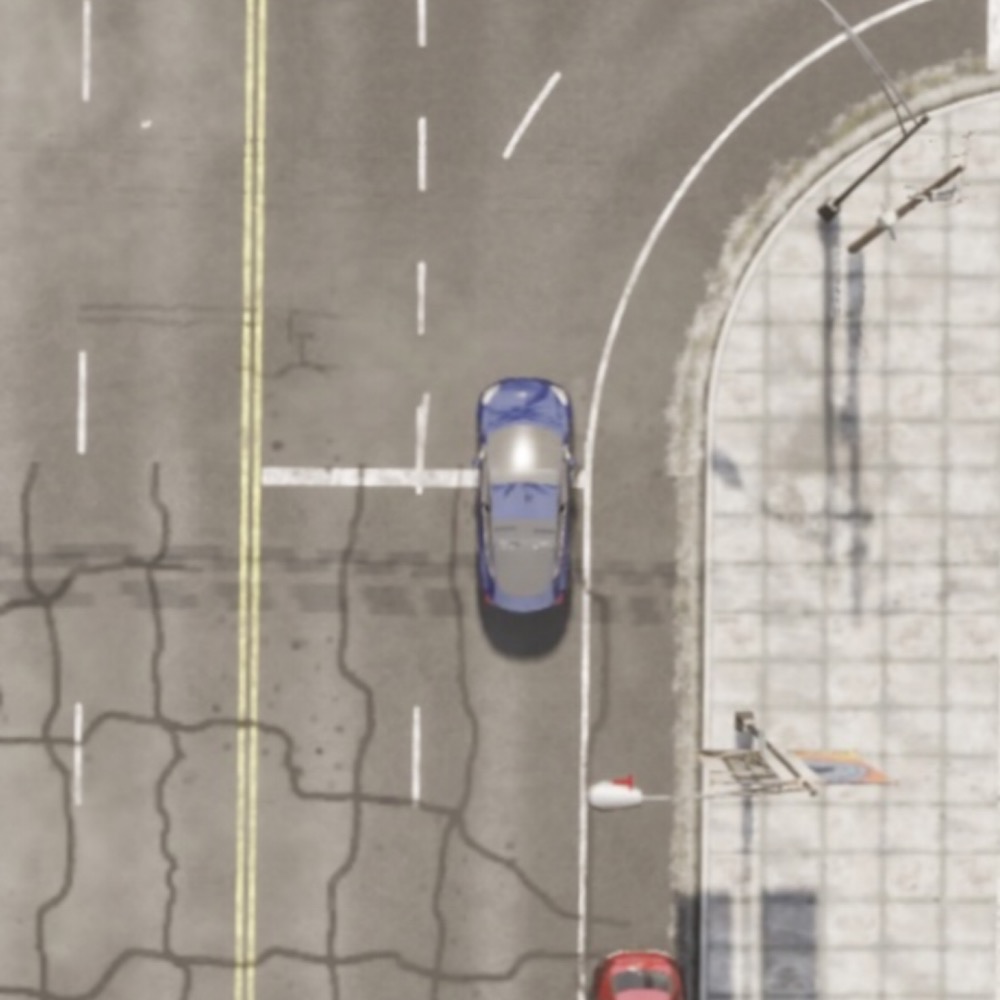}
        \caption{}
        \label{fig:bev4}
    \end{subfigure}
    \hfill
    \begin{subfigure}[b]{0.24\linewidth}
        \centering
        \includegraphics[width=\linewidth]{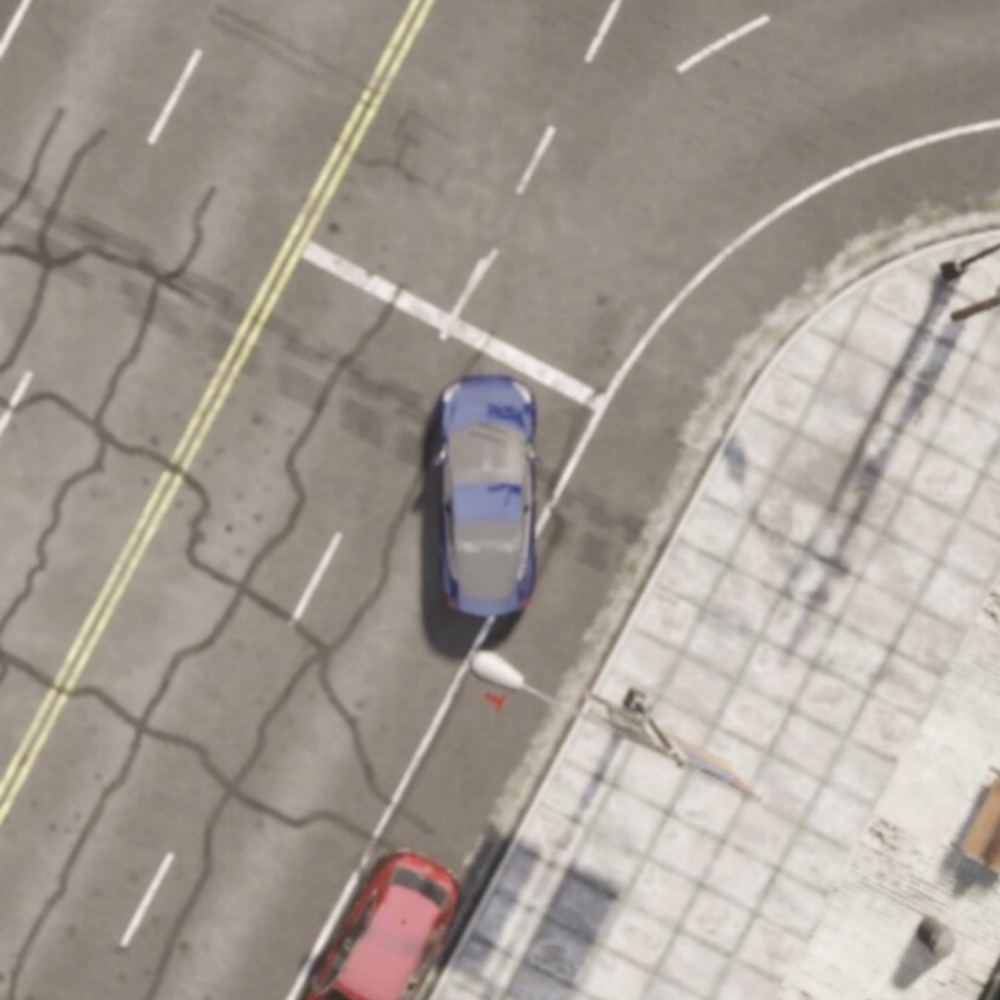}
        \caption{}
        \label{fig:bev5}
    \end{subfigure}
    \hfill
    \begin{subfigure}[b]{0.24\linewidth}
        \centering
        \includegraphics[width=\linewidth]{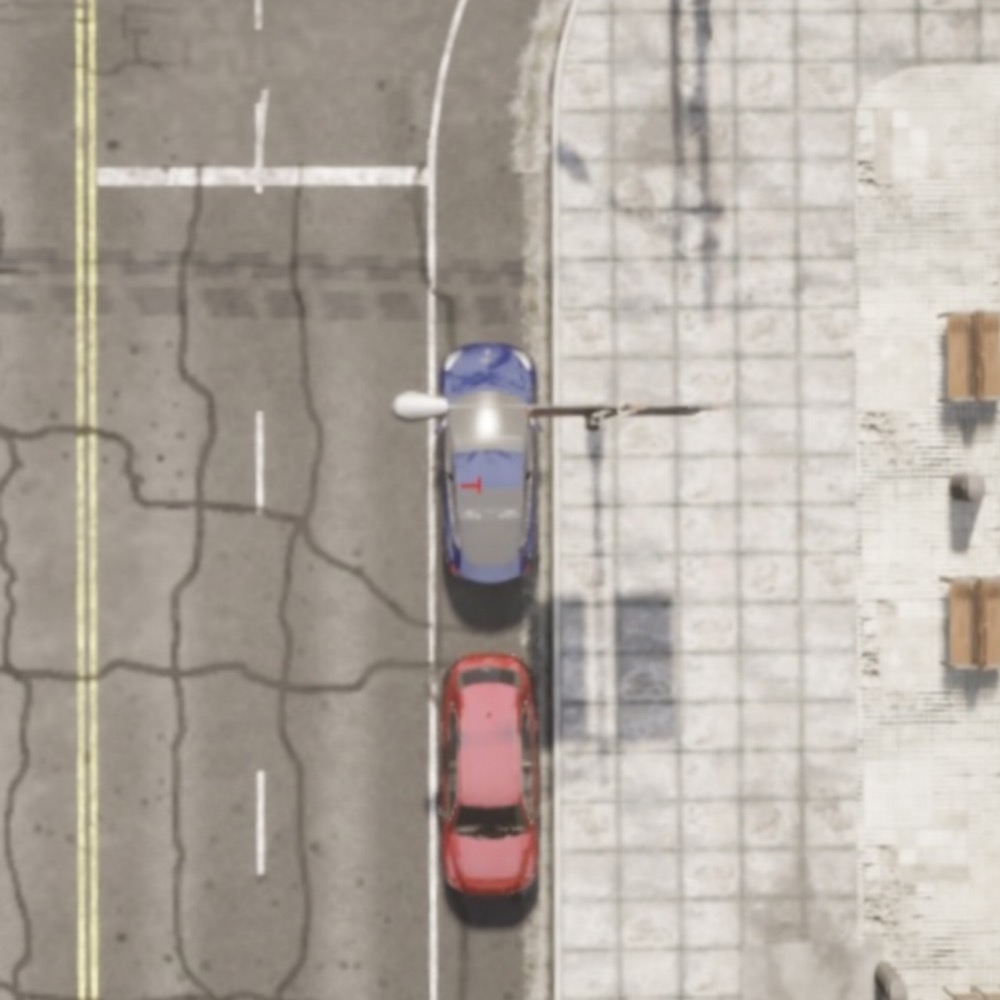}
        \caption{}
        \label{fig:bev6}
    \end{subfigure}
    \hfill
    \begin{subfigure}[b]{0.24\linewidth}
        \centering
        \includegraphics[width=\linewidth]{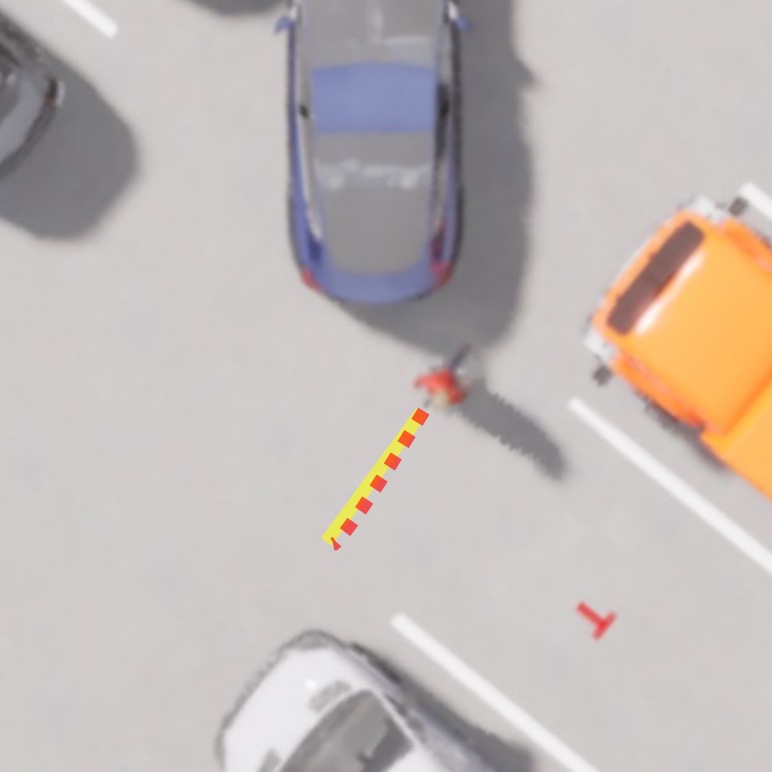}
        \caption{}
        \label{fig:ped}
    \end{subfigure}

    \caption{ (a)–(c) and (e)–(g) illustrate BEV views from Town04\_Opt and Town10HD\_Opt, with the red “T” indicating the target slot. In (a)–(c), the vehicle detects and yields to pedestrians. (d) shows the attention map of (a), highlighting goal-focused attention. (h) visualizes pedestrian prediction: yellow for ground-truth, red for predicted trajectory.}
    \label{fig:experience}
\end{figure}

\begin{table*}[t]
\caption{Comparative Evaluation of Parking Performance in Town04-Opt and Town10HD-Opt}
\label{tab:comparison_parking}
\centering
\begin{tabular}{lcccc|cccc}
\toprule
\multirow{2}{*}{\textbf{Method}} &
\multicolumn{4}{c|}{\textbf{Town04-Opt (Vertical Parking)}} &
\multicolumn{4}{c}{\textbf{Town10HD-Opt (Parallel Parking)}} \\
& \textbf{SR (\%) $\uparrow$} & \textbf{PE (m) $\downarrow$} & \textbf{OE (deg) $\downarrow$} & \textbf{CR (\%) $\downarrow$} 
& \textbf{SR (\%) } & \textbf{PE (m) } & \textbf{OE (deg) } & \textbf{CR (\%) } \\
\midrule
\textbf{Our approach}        & \textbf{96.57}  & \textbf{0.21}  & \textbf{0.41}  & \textbf{1.16}  & \textbf{97.30}  & \textbf{0.25}  & \textbf{0.43}  & \textbf{1.40} \\
E2E Parking~\cite{E2E Parking} & 91.41           & 0.30           & 0.87          & 2.08          & -           & -           & -          & - \\
\bottomrule
\end{tabular}
\end{table*}

\begin{table}[t]
\caption{Ablation Study on Town04\_HD}
\label{tab:ablation}
\centering
\setlength{\tabcolsep}{4pt}
\begin{tabular}{lcccc}
\toprule
\textbf{Method} & \textbf{SR (\%) } & \textbf{PE (m) } & \textbf{OE (deg) } & \textbf{CR (\%) } \\
\midrule
\textbf{baseline}                         & \textbf{96.57}  & \textbf{0.21}  & \textbf{0.41}  & \textbf{1.16} \\
w/o pedestrian                 & 78.36           & 0.23           & 0.43           & 20.54 \\
target concat         & 90.35           & 0.92           & 0.49           & 3.16 \\
\bottomrule
\end{tabular}
\end{table}

\subsection{Metrics}
To quantitatively assess the performance of the proposed autonomous parking model, we adopt the following evaluation metrics:

\textbf{Success Rate (SR)}: The proportion of trials in which the ego vehicle successfully parks in the designated slot without human intervention. This metric reflects the overall task completion capability.

\textbf{Position Error (PE)}: The Euclidean distance between the final position of the ego vehicle and the center of the target parking slot, computed for successful parking cases. It measures the spatial accuracy of the final parking pose.

\textbf{Orientation Error (OE)}: The absolute yaw angle difference between the ego vehicle and the target slot orientation in successful episodes. This metric evaluates the vehicle’s alignment precision.

\textbf{Collision Rate (CR)}: The probability of the ego vehicle experiencing any collision with static or dynamic objects during the parking process. It serves as an indicator of safety and robustness.

\subsection{Closed-loop Experiment}
To ensure the fairness of comparative evaluation, all models are trained under identical experimental settings, including dataset split, perception backbone, and optimizer hyperparameters. 
Figure~\ref{fig:experience} illustrates the parking behaviors of our method in Town04\_Opt and Town10HD\_Opt. The vehicle successfully parks in both vertical and parallel slot layouts. As shown in Figure~\ref{fig:att}, the attention maps focus on the target parking slot, enhancing interpretability of the decision process. Furthermore, Figures~\ref{fig:bev2} and~\ref{fig:ped} demonstrate the system's ability to predict pedestrian trajectories and take proactive action. When a potential collision is anticipated, the vehicle halts safely and resumes motion after the pedestrian has passed.

Table~\ref{tab:comparison_parking} summarizes the performance of our proposed method and the baseline E2E Parking~\cite{E2E Parking} across two representative parking scenarios. In the Town04-Opt (vertical parking) environment, our model outperforms the baseline across all metrics, including Success Rate (SR), Position Error (PE), Orientation Error (OE), and Collision Rate (CR). Specifically, PE and OE are reduced by approximately 30\% and 58.6\%, respectively, compared to E2E Parking. In the more challenging Town10HD-Opt (parallel parking) scenario, while E2E Parking does not report applicable results, our method achieves a high SR of 97.30\% and maintains a low CR of 1.40\%, demonstrating its robustness and generalization ability in complex real-world-like settings.

\subsection{Ablation Experiments}
\textbf{Pedestrian Prediction Module.} In Table~\ref{tab:ablation}, we evaluate the impact of removing the pedestrian prediction module. After this module is removed, the collision rate (CR) increases significantly, indicating that the absence of future pedestrian trajectory modeling severely impairs the system's ability to operate safely. The pedestrian prediction module, implemented with a GRU-based architecture, captures the temporal dynamics of pedestrian motion, allowing the vehicle to anticipate potential collisions rather than reacting passively. Without this foresight, the controller relies solely on current observations, leading to delayed or unsafe decisions when encountering moving obstacles. The results highlight that integrating predictive dynamics is essential for proactive planning and collision avoidance in complex traffic environments.

\textbf{Target Point Fusion Mechanism.} To assess the contribution of the target-guided fusion strategy, we replaced our proposed cross-attention-based mechanism with the simple position encoding concatenation strategy used in E2E Parking~\cite{E2E Parking}. As shown in Table~\ref{tab:ablation}, this substitution results in a noticeable increase in the L2 distance error, reflecting degraded localization accuracy for the parking slot. This performance drop stems from the insufficient integration between the target information and spatial features when using naive concatenation. In contrast, the cross-attention mechanism explicitly aligns BEV representations with the target slot, enabling the model to selectively emphasize goal-relevant spatial cues. Such dynamic feature modulation is critical for precise trajectory generation and consistent parking success.

\section{CONCLUSIONS}
In this paper, we present a Transformer-based end-to-end autonomous parking system designed for challenging and dynamic scenarios. The proposed network integrates BEV perception, goal-point attention fusion, and pedestrian trajectory modeling to directly predict control commands. By leveraging attention mechanisms and temporal modeling, our approach effectively captures spatiotemporal dependencies across multi-modal inputs. Extensive closed-loop experiments conducted in the CARLA simulation environment demonstrate that our method significantly outperforms existing end-to-end baselines in both accuracy and robustness.  

Currently, our framework is validated in simulation, and its generalization to real-world conditions remains to be explored. In future work, we aim to deploy the system in real vehicles, extend it to diagonal and oblique parking scenarios, and incorporate additional sensors such as fisheye cameras and ultrasonic radars. Furthermore, integrating reinforcement learning will be explored to enhance adaptability and collision avoidance in complex traffic environments.




\end{document}